\documentclass[12pt]{article}
\usepackage[T1]{fontenc} 
\fontfamily{bch}\selectfont
\usepackage[skip=8pt plus1pt, indent=0pt]{parskip}
\usepackage{titlesec}
\usepackage{fancyhdr}
\usepackage{setspace}
\usepackage{lmodern}
\usepackage{caption}
\usepackage{subcaption}
\usepackage[none]{hyphenat}

\newlength\titleindent
\newlength\sectionskip
\titleformat{\section}  
  {\vspace{\sectionskip}\fontsize{24}{24}\sffamily\raggedright} 
  {\llap{\parbox[b]{\titleindent}{\thesection\hfill}}} 
  {0em} 
  {} 
  [] 
\titlespacing*{\section}{0pt}{0pt}{24pt plus 1pt}

\titleformat{\subsection}  
  {\fontsize{15}{15}\sffamily\bfseries\raggedright} 
  {\thesubsection} 
  {0.6em} 
  {} 
  [] 
\titlespacing*{\subsection}{0pt}{18pt plus 3pt minus 1pt}{10pt plus 1pt}

\titleformat{\subsubsection}  
  {\fontsize{13}{13}\sffamily\bfseries\raggedright} 
  {\thesubsubsection} 
  {0.6em} 
  {} 
  [] 
\titlespacing*{\subsection}{0pt}{24pt plus 3pt minus 1pt}{12pt plus 1pt}

\usepackage[headheight=15pt]{geometry}  
\usepackage[intlimits]{mathtools}       
\usepackage{graphicx}                   
\usepackage{booktabs}                   
\usepackage{multirow}                   
\usepackage[normalem]{ulem}             
\useunder{\uline}{\ul}{}                
\usepackage{appendix}                   
\usepackage{xspace}
\usepackage{amsthm}                     
\usepackage{hyperref}                   
\usepackage{natbib}                     
\usepackage{lscape}                     
\hypersetup{
    pdftitle={Target Population Synthesis using CT-GAN},
    pdfauthor={Tanay Rastogi},
    colorlinks=true,
    linkcolor=black,
    citecolor=black,
    filecolor=black,
    urlcolor=black
}

\begin{document}

\title{Workplace Location Choice Model based on Deep Neural Network}
\author{Tanay Rastogi and Anders Karlström}
\date{}
\maketitle
\pagenumbering{arabic}

\begin{abstract}
\noindent Discrete choice models (DCMs) have long been used to analyze workplace location decisions, but they face challenges in accurately mirroring individual decision-making processes. This paper presents a deep neural network (DNN) method for modeling workplace location choices, which aims to better understand complex decision patterns and provides better results than traditional discrete choice models (DCMs). The study demonstrates that DNNs show significant potential as a robust alternative to DCMs in this domain. While both models effectively replicate the impact of job opportunities on workplace location choices, the DNN outperforms the DCM in certain aspects. However, the DCM better aligns with data when assessing the influence of individual attributes on workplace distance. Notably, DCMs excel at shorter distances, while DNNs perform comparably to both data and DCMs for longer distances. These findings underscore the importance of selecting the appropriate model based on specific application requirements in workplace location choice analysis.
\end{abstract}


\newpage
\section{Introduction}\label{Introduction}
Integrated Land-Use and Transportation Models (ILUTMs) are robust analytical frameworks that merge land use and transportation systems to capture the dynamic interplay between residential, employment, and commuting patterns. As highlighted by \cite{Levine1998RethinkingBalance, Wang2011IncrementalModeling}, a key feature of these models is the workplace location choice component, which links job location decisions within the land-use model to travel behaviors in the transportation model. This connection makes sure that when one system changes, the other one does too, allowing ILUTMs to predict commuting habits and travel needs accurately—this is crucial for creating effective transportation networks. Additionally, the workplace location choice model helps evaluate policies that affect where jobs are located, like incentives for businesses or investments in transportation infrastructure. For example, the introduction of a new subway line could render certain areas more attractive for businesses, thereby affecting both job creation and commuting patterns.

Discrete choice models (DCMs) are frequently employed to analyze workplace location decisions, offering valuable insights into spatial decision-making processes. These models account for factors such as accessibility, suitability, and spatial competition. For instance, \cite{Inoa2015EffectEmployment} utilized a three-level nested logit model to examine the interdependent choices of residential location, workplace, and employment type. Their findings indicated that individual-specific accessibility to employment is a critical determinant of residential location, and that the attractiveness of various employment types is a stronger predictor of workplace location than the sheer number of available jobs. Similarly, \cite{Ho2016AEffects} used multinomial logit and nested logit models to find out how both the clustering of jobs and competition between locations affect where people choose to work, with competition being the more important factor. Additionally, \cite{Jiao2015JointBehavior} created and tested different mixed logit models for where households choose to live and work, finding that a combined model—considering how these choices affect each other—is the most precise, with travel distance and access to rail transit being the most important factors. Furthermore, \cite{Vitins2016IntegrationSingapore} introduced a workplace choice model that incorporates destination capacity constraints, applying it within a large-scale transportation simulation in Singapore that leverages recent computational advances. Likewise, \cite{Naqavi2023MobilityStockholm} proposed a workplace accessibility measure that integrates individuals’ constraints in time, space, and resources to evaluate the impact of non-marginal land use policy changes, employing a 2-level nested logit model based on an activity framework to render workplace location an endogenous component of the decision process.

A central challenge in developing DCMs is formulating a specification that accurately mirrors an individual’s decision-making process—especially with regard to variable selection and representation. Building a DCM requires assumptions about decision-makers’ rules, information-processing strategies, consideration sets, and utility functions. Research by \cite{Torres2011HowExperiments} demonstrates that properly capturing systematic taste heterogeneity can be particularly challenging, especially when dealing with novel or complex decision problems without prior knowledge. Mis-specification of utility functions may lead to diminished predictability, biased estimates, and ultimately misguided policy decisions. As noted by \cite{Ben-Akiva2002HybridChallenges}, even advanced DCMs such as Mixed Logit and Latent Class Choice Models—which attempt to address taste heterogeneity—still depend on a priori knowledge of both the systematic utility and the random error structure. Merely accounting for random heterogeneity does not fully eliminate the bias introduced by incorrectly specified systematic utilities.

In recent years, deep neural networks (DNNs) have emerged as a popular alternative for modeling choice behavior. DNNs are capable of approximating virtually any function directly from data, thereby learning flexible representations from large datasets—capabilities that often exceed those attainable with hand-engineered features crafted by domain experts. The concept of substituting traditional logit models with neural networks was pioneered by \cite{Bentz2000NeuralApproach}, who employed a simple feed-forward network for market analysis. Their work demonstrated that DNN-based models can reveal the underlying structure of a problem directly from the data, without relying on pre-established theoretical assumptions about the data-generating process. This data-driven approach not only streamlines model selection but also reduces the influence of subjective biases. Moreover, learning directly from data provides the added advantage of uncovering unexpected patterns. For example, \cite{vanCranenburgh2022ChoicePaper} offers a comprehensive review of various DNN-based models applied to choice modeling tasks—ranging from travel mode and vehicle type to train type—and compares them to traditional DCMs such as the Multinomial Logit (MNL), Nested Logit (NL), and Latent Class MNL models. In a related effort, \cite{Kim2024AModel} introduced a model based on a Lattice Network (LN), termed DCM-LN, which was evaluated using both Monte Carlo simulations and Swiss Metro data to estimate willingness to pay (WTP). Their approach balances flexibility and interpretability by enforcing monotonicity in the utility function for selected attributes while capturing complex non-linear and interaction effects in a data-driven manner through lattice networks. Additionally, \cite{Nam2020DeepBehavior} proposed a DNN model for travel mode choice prediction that outperforms traditional discrete choice models in prediction accuracy at both individual and aggregate levels. Similarly, \cite{Wang2021AnBike} developed an attention-based deep learning framework for predicting the trip destinations of bike-sharing users, achieving superior performance compared to DCM-based approaches on real-world datasets. Further extending these advances, \cite{Wang2024DeepBehavior} presented a novel deep learning model for short-term origin-destination distribution prediction in urban rail transit networks by integrating a destination choice component with a deep learning module to capture both behavioral and spatio-temporal dynamics.

The studies discussed above—as well as the survey by \cite{vanCranenburgh2022ChoicePaper}—primarily focus on the application of DNNs in discrete choice scenarios with relatively small choice sets (typically 5–10 alternatives), such as travel mode selection. In contrast, workplace location choice involves a substantially larger number of alternatives. Traditional DCMs often address this challenge by sampling from the set of alternatives, as described by \cite{Nerella2004NumericalModels}; however, the extensive choice set complicates the formulation of an optimal model specification. By contrast, DNNs, with their inherent ability to learn complex structures from data, are well-suited to modeling workplace location choices. To the best of the author's knowledge, no studies have yet explored the application of DNNs specifically for workplace location choice. To address this gap, the present study proposes the application of DNNs to predict workplace location choices.

The remainder of the article is organized as follows: Section \ref{Choice Model} introduces the workplace location choice models used in this study. We begin with the DCM-based model and then present the proposed DNN-based models for predicting workplace location choices. The models are evaluated using travel survey data from Stockholm and SAMS-based data from Statistics Sweden. Section \ref{Experiment} details the experimental setup and model training results. Following this, Section \ref{Compare} offers a comprehensive comparison between the DCM and DNN models across various attributes. Finally, Section \ref{Conclusion} concludes the article by summarizing the analysis and suggesting potential directions for future research.

\section{Workplace Choice Models}\label{Choice Model}
In this section, we describe the DCM-based and DNN-based models used in this study. The DCM serves as a benchmark for evaluating the performance of the DNN models. We propose two distinct DNN models, which differ in their input data. The first DNN model replicates the input of the DCM, while the second model utilizes all available data as input. 

\subsection{Discrete Choice Model (DCM)}
In this study, we use a simple 2-level Nested Logit (NL) model proposed by \cite{Naqavi2023MobilityStockholm} as the DCM-based workplace location choice model. In their research, they introduces the spare time accessibility measure based on the dynamic activity-based travel demand model. The spare time accessibility for an individual living in zone $i$ and going to work in zone $j$, is given by the expected maximum utility of an individual’s spare time activity-travel patterns conditional on work zone and home zone. This measure considers temporal and spatial constraints, socioeconomic characteristics of the individual, activity participation as well as travel mode and travel time. 

\begin{figure}[ht]
\centering
{{\includegraphics[scale=0.6]{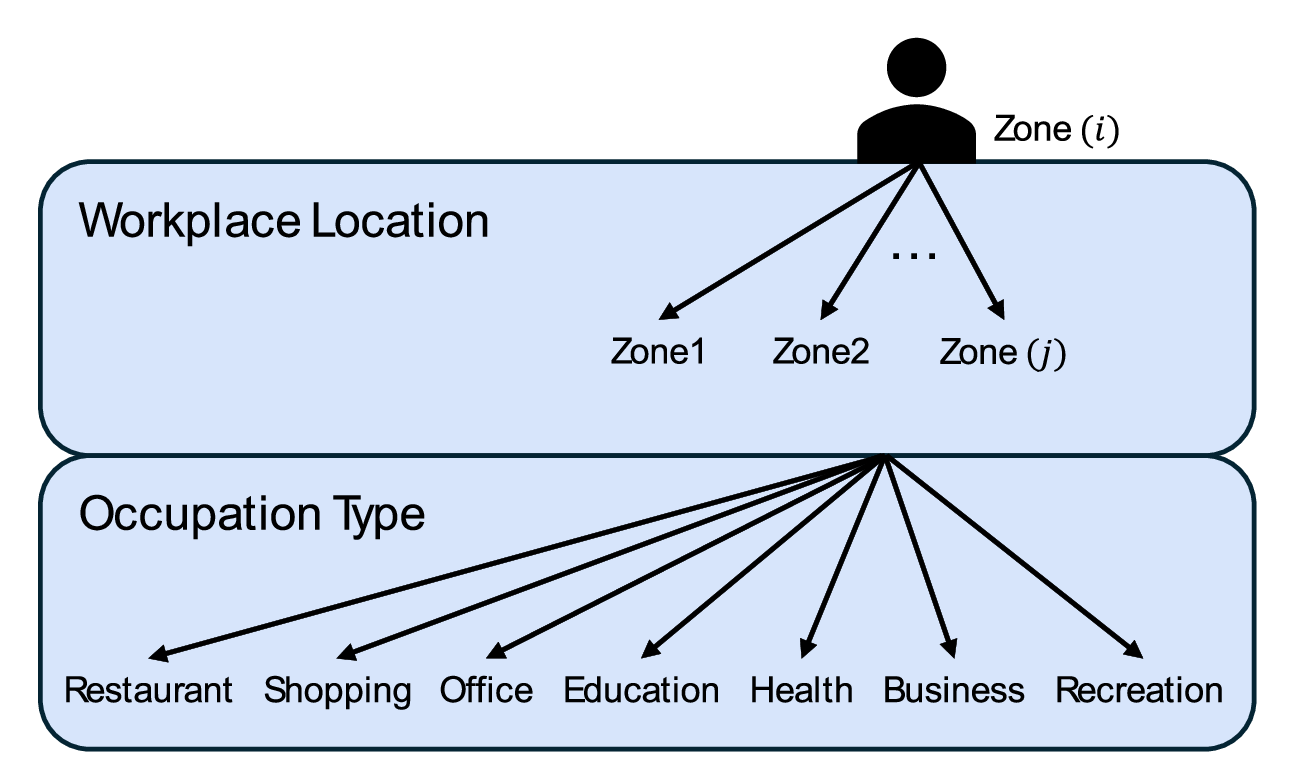}}}
\caption{2-level NL model for workplace location choice model proposed by \cite{Naqavi2023MobilityStockholm}.}
\label{fig:fatemeh_model}
\end{figure}

Figure \ref{fig:fatemeh_model} illustrates the NL model's structure, where an individual's choice of a specific workplace depends on: (1) the workplace location, defined by the zone where the workplace is situated; (2) the occupation, which includes different types of jobs available to the individual in each zone; (3) characteristics of the specific individual; and (4) spare time accessibility conditional on the home and workplace. Since occupational and workplace choices are unobserved in our data, the model is aggregated to the workplace location level.

\newpage
The systematic utility,  $v_{n, ij}$ of the NL workplace location choice model for individual $n$, choosing a workplace $j$ which is conditional on home zone $i$, is give as, 
\begin{align}
\label{eq:nl_utils}
v_{n, ij} = (\beta_A + \beta_{Acr}1_n(cr))A_{n, ij} + \lambda \log\sum_{k=1}^{K}e^{\alpha_{k}/\lambda+\log(N_{jk})}
\end{align} 

Here, $\beta_A$ is the parameter for the spare time accessibility measure $A_{n, ij}$, individual characteristics is represented by $1_n(cr)$ as a dummy variable indicating if the individual has access to personal car with a parameter $\beta_{Acr}$, $\lambda$ is the structural log-sum parameter of the NL model, and $\alpha_{k}$ represents occupation-specific constants for different types of occupations. The log-sum term represents the expected utility of choosing any of the workplaces available in zone $j$, where these workplaces belong to an occupational type $(k \in 1, ...K)$ and there are $N_{jk}$ such workplaces. The nested logit model is integrated with an activity-based model called SCAPER, developed by \cite{Vastberg2020AModel}, which is used to derive the spare time accessibility employed in the model.

The probability of observing a workplace location choice, given home zone $i$, is expressed as:
\begin{align}
\label{eq:nl_prob}
P_{n, j|i} = \frac{e^{v_{n, ij}}}{\sum_{j=1}^{J}e^{v_{n, ij}}}
\end{align} 

The model parameters, $(\alpha, \beta, \lambda)$ are estimated using a maximum likelihood approach, where the likelihood function is formulated based on the above probabilities. The likelihood function is given by:
\begin{align}
\label{eq:loss}
LL(\alpha, \beta, \lambda) = -w_n\left( \sum_{n=1}^{N}\sum_{j}^{} 1_{y_{nj}}\ln\left( \Pr(j|i) \right) \right)
\end{align} 
where $1_{y_{nj}}$ is boolean for the chosen workplace and $w_n$ is the sampling weight for the individual $n$.

\subsection{Deep Neural Network (DNN)}
The proposed DNN-based workplace location choice model is designed to approximate the NL-based DCM model presented earlier. As described in the previous section, the workplace location choice model involves an individual selecting a workplace location $j$, conditional on home zone $i$, from a zone choice set $Z$ where $\forall i,j \in Z$. The input to the model includes observed data for each zone, such as the number of available jobs for each occupational type $N_{jk}$ across the different occupational categories $(k \in 1, ...K)$, the personal characteristics of the individual $z_n$, and the spare time accessibility measure $A_{n, ij}$ calculated for each individual between their work zone and home zone.

\begin{figure}[ht]
\centering
{{\includegraphics[scale=0.86]{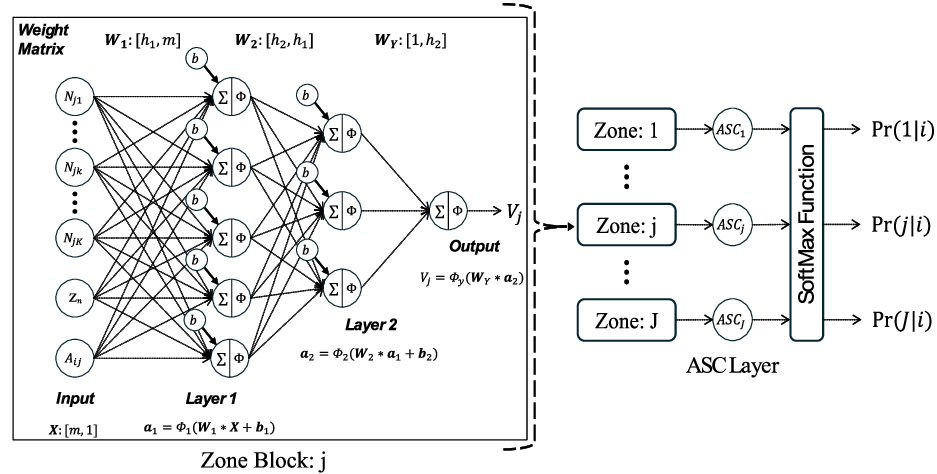}}}
\caption{Structure of the proposed DNN for workplace location choice model.}
\label{fig:DNN}
\end{figure}

The proposed DNN-based model consists of two main components: a zone block and an alternative-specific constant (ASC) layer as presented in the Figure \ref{fig:DNN}. The zone block is designed as a fully-connected multi-layer feed-forward network to compute the systematic utilities for each zone $j$, conditional on the home zone $i$. This zone block comprises three or more layers, including an input layer, one or more hidden layers, and an output layer with nonlinearly-activating nodes. In the MLP, each layer is fully connected to the subsequent layer, with information flowing in one direction—from input to output—ensuring no cycles or loops in the network architecture. 

The zone block takes as input the number of jobs by occupational type, individual-specific characteristics, and the spare time accessibility measure, represented collectively as $\textbf{X}$, a vector of dimension $[m, 1]$. The output is a single value representing the utility of zone block $j$, denoted as $V_j$.

For an MLP with layers $l = 1, 2 .. k-1$, where each layer $l$ contains $h_l$ neurons, the layer computations are defined as:
\begin{align}
a_l = \phi(W_l \cdot a_{l-1} + b_l)
\label{eq:dnn_forward}
\end{align}

The final output of the zone block, representing the systematic utility for zone $j$, is obtained at layer $l=k$:
\begin{align}
V_j = \phi(W_k \cdot a_{k})
\label{eq:util}
\end{align}

In these equations, the matrix $W_l$ represents the weights of layer $l$, with dimensions $[h_l, h_{l-1}]$, and the vector $b_l$ represents the biases for layer $l$, with dimensions $h_l$. For the input layer $l=0$, we have $a=\textbf{X}$, and the weight matrix dimensions are $[h_l, m]$. The non-linear function $\phi$ serves as the activation function, which modifies values between layers. In the proposed model, the rectified linear unit (ReLU) activation function is used for all layers in the zone block.

In the later stages of the network, the output of each zone block is combined with a trainable parameter, referred to as the ASC, which represents the alternative specific constant for that particular zone block. As a result, the proposed model includes $J$ ASC parameters, one for each zone block. Hence the resultant systematic utility after the ASC layer is calculated as, 
\begin{align}
\textbf{V} = V_j + ASC_j \; \forall j=1, .., J
\label{eq:ASC}
\end{align}

Finally, similar to DCM-based models, the probability $Pr(j|i)$ of observing a workplace location $j$, given home zone $i$, is expressed as:
\begin{align}
\Pr(j|i) = softmax(\textbf{V}) = \frac{e^{V_{j}}}{\sum_{j=1}^{J}e^{V_{j}}}
\label{eq:Pr}
\end{align}

The weights and biases of the DNN-based model are estimated using the backpropagation training method with the Adam optimizer. The model is trained using a mini-batch approach, where the optimizer minimizes the likelihood function presented in Equation \ref{eq:loss}. Furthermore, the model necessitates fine-tuning of its hyperparameters, which include the number of hidden layers, the number of neurons in each hidden layer, the learning rate for the optimizer, and the number of epochs for training.

\section{Data and Estimation Results}\label{Experiment}

\subsection{Data}
The data for this study comes from two distinct sources: the Stockholm travel survey conducted in 2015 and SAMS-based data provided by Statistics Sweden. The travel survey offers detailed full-day travel diaries for individuals who worked on weekdays. It includes information on 6,204 individuals selected through a stratified sampling process from the Swedish total population register. This dataset captures their socioeconomic characteristics, such as their residential zone and workplace zone. Table \ref{tab:indattrib} outlines the individual-specific attributes from the travel survey dataset that were used in this study.

\begin{table}[htbp]
\centering
\caption{Summary of the individual specific attributes from travel survey.}
\label{tab:indattrib}
\resizebox{0.8\textwidth}{!}{%
\begin{tabular}{@{}clc@{}}
\textbf{Attributes} & \multicolumn{1}{c}{\textbf{Category}}                          & \textbf{Count} \\ \midrule
household type      & (1) Single parent, no kids                                     & 1072           \\
                    & (2) Multi parent, no kids                                      & 2441           \\
                    & (3) Single parent, kids \textbackslash{}textgreater 10 years   & 102            \\
                    & (4) Multi parent, kids \textbackslash{}textgreater 10 years    & 768            \\
                    & (5) Single parent, kids \textbackslash{}textless\{\}= 10 years & 92             \\
                    & (6) Multi parent, kids \textbackslash{}textless\{\}= 10 years  & 1729           \\
has kids?           & (0) No                                                         & 3513           \\
                    & (1) Yes                                                        & 2691           \\
has car?            & (0) No                                                         & 1998           \\
                    & (1) Yes                                                        & 4206           \\
gender              & (0) Male                                                       & 2781           \\
                    & (1) Female                                                     & 3423           \\
household income    & (1) 0 - 10,000                                                 & 12             \\
                    & (2) 10,001 - 13,000                                            & 28             \\
                    & (3) 13,001 - 17,000                                            & 43             \\
                    & (4) 17,001 - 22,000                                            & 115            \\
                    & (5) 22,001 - 28,000                                            & 319            \\
                    & (6) 28,001 - 36,000                                            & 557            \\
                    & (7) 36,001 - 51,000                                            & 954            \\
                    & (8) 51,001 - 80,000                                            & 2154           \\
                    & (9) 80,001 - 120,000                                           & 1378           \\
                    & (10) 120,001 - 170,000                                         & 405            \\
                    & (11) 170,001 and above                                         & 239            \\
employment          & (1) full-time                                                  & 5350           \\
                    & (2) part-time                                                  & 818            \\
                    & (3) not-working                                                & 29             \\
                    & (4) No data                                                    & 7              \\ \midrule
\textbf{Training}   & \multicolumn{2}{c}{4653}                                                        \\
\textbf{Validation} & \multicolumn{2}{c}{1551}                                                        \\ \bottomrule
\end{tabular}%
}
\end{table}

\begin{figure}[htbp]
\centering
{{\includegraphics[scale=0.9]{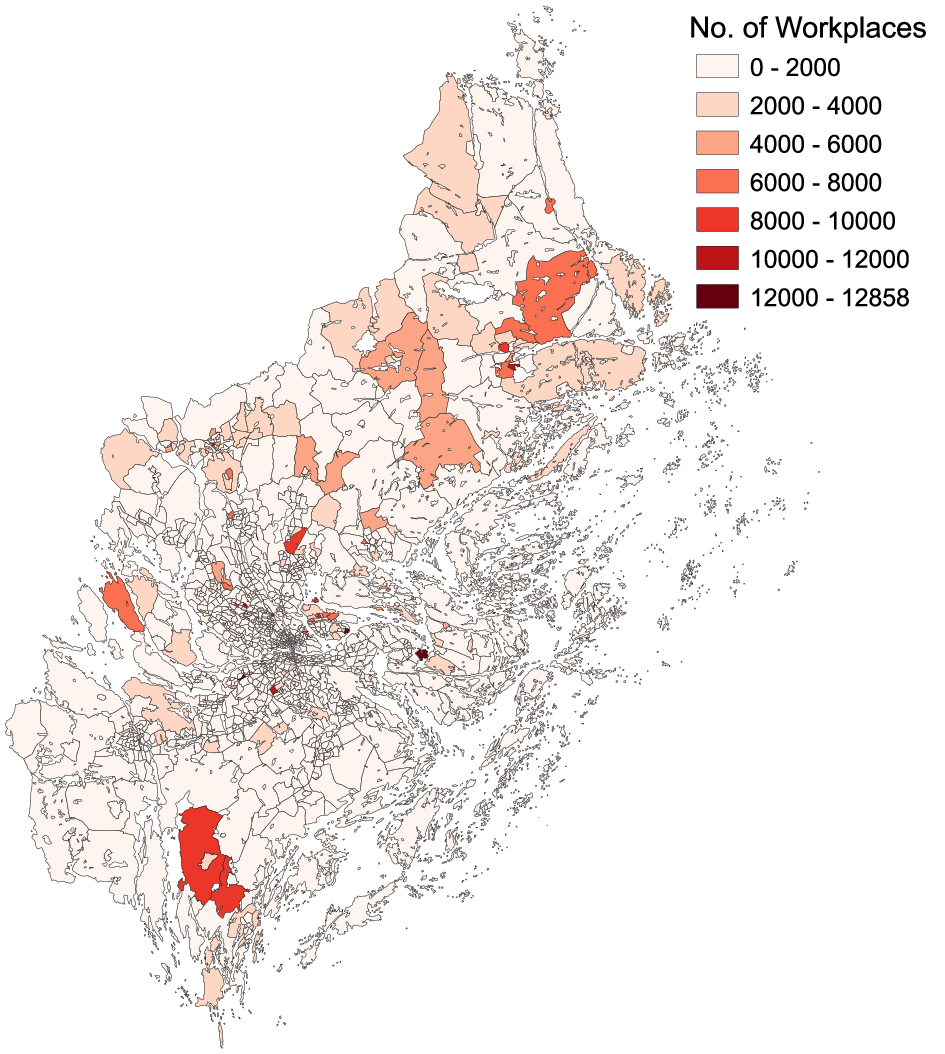}}}
\caption{Total number of workplaces in each of 1375 zones in Stockholm.}
\label{fig:totWrk}
\end{figure}

The second dataset uses SAMS-based data from Statistics Sweden to report the number of workplaces per occupational type in each zone. In our model, we divided the Stockholm region into 1,375 zones based on SAMPERS\footnote{For a report on the origins of SAMPERS see \cite{Beser2002SAMPERSTool}}, aggregating the SAMS data accordingly. From this dataset, we extracted the number of workplaces per zone for various employment types, including (1) restaurants, (2) shopping, (3) offices, (4) education, (5) health, (6) business, and (7) recreation. Figure \ref{fig:totWrk} illustrates the distribution of total available workplaces across the 1,375 zones in Stockholm.

Additionally, spare time accessibility is calculated using the SCAPER model, as presented by \cite{Vastberg2020AModel}, and is used as an input in the workplace location choice model. SCAPER provides spare time accessibility for all sampled individuals from their home zone to all possible work zones in Stockholm, denoted as $ A_{n, ij} $.

To estimate the parameters for both the DCM and DNN models, the dataset was divided into training and validation sets. The models were trained on the training set and subsequently evaluated using the unseen validation set to ensure they did not overfit the training data. The travel survey data was split in a 75/25 ratio, with 4653 randomly sampled individuals in the training set and 1551 in the validation set. The spare time accessibility data was also split in the same ratio, corresponding to the individuals in each set.

\subsection{Training Results}\label{Results}

\subsubsection{DCM}
The DCM-based NL model was initially estimated on the training dataset by maximizing the log-likelihood based on the probabilities of individuals’ workplace location choices, as defined by Eq. \ref{eq:nl_utils}–\ref{eq:loss}. The model parameters were estimated using the unconstrained quasi-Newton optimization methods available in MATLAB2022a. The estimated model incorporates size attributes for different occupational types, corresponding occupational-specific constants, a log-sum parameter, the spare time accessibility measure, and its interaction with a dummy variable indicating access to a personal vehicle. Table \ref{tab:DCM-res} presents the estimated parameters, along with the final likelihood values for both the training and validation sets, calculated as -32827.40 and -11224.46, respectively. 

\begin{table}[htbp]
\centering
\caption{Parameter estimated for discrete choice NL workplace choice model.}
\label{tab:DCM-res}
\resizebox{\textwidth}{!}{%
\begin{tabular}{@{}llcccc@{}}
\multicolumn{1}{c}{\textbf{Parameter}} &
  \multicolumn{1}{c}{\textbf{Attribute}} &
  \textbf{Estimate} &
  \textbf{Std} &
  \textbf{T-value} &
  \textbf{T-value against 1} \\ \midrule
$\alpha_{1}$  & Restaurant                            & 1.038               & 0.023 & 45.880 &        \\
$\alpha_{2}$  & Shopping                              & 0.983               & 0.012 & 78.728 &        \\
$\alpha_{3}$  & Office                                & 0.867               & 0.019 & 46.469 &        \\
$\alpha_{4}$  & Eduation                              & 1.009               & 0.014 & 74.714 &        \\
$\alpha_{5}$  & Health                                & 0.965               & 0.012 & 83.104 &        \\
$\alpha_{6}$  & Business                              & 0.893               & 0.011 & 83.806 &        \\
              & Recreation (ref)                      & -                   & -     & -      &        \\
$\lambda$     & log-sum (Occupations)                 & 1.048               & 0.019 & 54.427 & -2.470 \\
$\beta_{A}$   & $A_{ij}$                              & 0.567               & 0.015 & 37.867 &        \\
$\beta_{Acr}$ & $A_{ij}*has-car$                      & -0.128              & 0.017 & -7.341 &        \\ \midrule
              & \multicolumn{1}{c}{\textbf{Training}} & \textbf{Validation} &       &        &        \\ \midrule
$LL(\beta)$ &
  \multicolumn{1}{c}{-32827.40} &
  -11224.46 &
  \multicolumn{1}{l}{} &
  \multicolumn{1}{l}{} &
  \multicolumn{1}{l}{} \\
$LL(0)$ &
  \multicolumn{1}{c}{-37261.89} &
  - &
  \multicolumn{1}{l}{} &
  \multicolumn{1}{l}{} &
  \multicolumn{1}{l}{} \\
\# observations &
  \multicolumn{1}{c}{4653} &
  1551 &
  \multicolumn{1}{l}{} &
  \multicolumn{1}{l}{} &
  \multicolumn{1}{l}{} \\ \bottomrule
\end{tabular}%
}
\end{table}

\subsubsection{DNN}
To validate our proposed model, we trained two DNN models that differed solely by their input values. The first model uses only the "has\_car" attribute—similar to the DCM model—and is referred to as DNN-Car. The second model uses all available individual attributes as input, thereby increasing the dimension of its input vector; this model is referred to as DNN-All. Both models were trained and tested on NVIDIA GeForce RTX 3080 units (each equipped with 8GB of memory) for a total of 200 epochs. For both DNN models, we optimized the hyperparameters—including the number of layers, the number of neurons per layer, and the learning rate for the Adam optimizer. The best hyperparameter sets, along with their corresponding log likelihood values for both the training and validation sets, are presented in Table \ref{tab:DNN-res}.

\begin{table}[htbp]
\centering
\caption{Hyper-parameters and final log likelihood the trained DNN models.}
\label{tab:DNN-res}
\resizebox{0.8\textwidth}{!}{%
\begin{tabular}{@{}lcccc@{}}
                & \multicolumn{2}{c}{\textbf{DNN-Car}}    & \multicolumn{2}{c}{\textbf{DNN-All}}    \\ \midrule
\multicolumn{1}{c}{\textbf{Hyper-params}} & \multicolumn{2}{c}{\textbf{Value}} & \multicolumn{2}{c}{\textbf{Value}} \\ \midrule
N Inputs        & \multicolumn{2}{c}{9}                   & \multicolumn{2}{c}{15}                  \\
N Layers        & \multicolumn{2}{c}{3}                   & \multicolumn{2}{c}{3}                   \\
N Neurons       & \multicolumn{2}{c}{{[}100, 150{]}}      & \multicolumn{2}{c}{{[}100, 150{]}}      \\
Learning Rate   & \multicolumn{2}{c}{0.01}                & \multicolumn{2}{c}{0.01}                \\
Epochs          & \multicolumn{2}{c}{200}                 & \multicolumn{2}{c}{200}                 \\ \midrule
                & \textbf{Training} & \textbf{Validation} & \textbf{Training} & \textbf{Validation} \\ \midrule
$LL(\beta)$     & -30926.56         & -10993.44           & -30480.58         & -10943.76           \\
$LL(0)$         & -38425.63         & -                   & -38425.63         & -                   \\
\# observations & 4653              & 1551                & 4653              & 1551                \\ \bottomrule
\end{tabular}%
}
\end{table}

\section{Model Comparison}\label{Compare}
The training results presented in Section \ref{Results} reveal a significant finding: both DNN models outperform the DCM-based approach, as shown by their lower average likelihood values across the training and validation datasets (see Table \ref{tab:results}). This outcome indicates that the DNN models offer a more precise representation of the observed choice preferences. Moreover, the DNN-All model—which utilizes all available individual attributes—outperforms the DNN-Car model, emphasizing the importance of individual characteristics in workplace choice decisions, a factor omitted in the other model.

\begin{table}[htbp]
\centering
\caption{Average loglikelihood for the DCM and DNN models on both training and validation data.}
\label{tab:results}
\resizebox{0.7\textwidth}{!}{%
\begin{tabular}{@{}lcccc@{}}
                    & \textbf{\#Obs.} & \textbf{DCM} & \textbf{DNN-Car} & \textbf{DNN-All} \\ \midrule
\textbf{Training}   & 4653            & -7,055       & -6,647           & -6,551           \\
\textbf{Validation} & 1551            & -7,237       & -7,088           & -7,056           \\ \bottomrule
\end{tabular}%
}
\end{table}

To further evaluate these models, we used the validation dataset to conduct a comprehensive analysis using additional quantitative metrics. In the subsequent sections, we examine in detail the impact of both workplace attributes and individual characteristics, aiming to shed light on the nuanced ways different variables contribute to the decision-making process.

\subsection{Workplace Attributes}
In this study, each zone is characterized by seven types of workplaces: (1) restaurants, (2) shopping, (3) offices, (4) education, (5) health, (6) business, and (7) recreation. A common trend in workplace location choice is that individuals tend to select areas with more job opportunities, and their decisions are also influenced by the specific types of jobs available. To quantify the relationship between individual choices and the availability of various workplace types, we calculate the Pearson correlation coefficient between workplace attributes and individual workplace choices.

The Pearson correlation coefficient \citep{Berman2016UnderstandingData} quantifies the linear relationship between two vectors, with values ranging from -1 to +1. A value of 0 indicates no correlation, while -1 or +1 signifies an exact linear relationship. Positive values imply that as one variable increases, the other also increases; negative values indicate that as one variable increases, the other decreases. The coefficient between two vectors, $x$ and $y$, is calculated as follows:

\begin{align}
\label{eq:pearson_coff}
r = \frac{\sum (x_i - \bar{x})(y_i - \bar{y})}{\sqrt{\sum (x_i - \bar{x})^2} \sqrt{\sum (y_i - \bar{y})^2}}
\end{align}

Table \ref{tab:pearson-coff} presents the statistics and two-tailed p-values for the correlations between the number of various workplace types and the frequency with which individuals in the validation dataset choose each zone. The results clearly demonstrate that the number of work opportunities in a zone significantly influences individual choices, as indicated by the high positive correlation values observed in both the validation dataset and model outputs. Comparing the three models with the validation data further reveals that the DNN models align more closely with the observed data than the DCM-based model. This finding suggests that the DNN models more effectively capture the variability in workplace choice resulting from changes in the number of work opportunities.

\begin{table}[htbp]
\centering
\caption{Pearson correlation coefficient and p-value between workplace attributes and workplace chosen by individuals. Stats closest to the validation are in bold.}
\label{tab:pearson-coff}
\resizebox{\textwidth}{!}{%
\begin{tabular}{@{}lllllllll@{}}
 &
  \multicolumn{2}{c}{\textbf{Validation Data}} &
  \multicolumn{2}{c}{\textbf{DCM}} &
  \multicolumn{2}{c}{\textbf{DNN-Car}} &
  \multicolumn{2}{c}{\textbf{DNN-All}} \\ \midrule
\multicolumn{1}{c}{\textbf{Attribute}} &
  \multicolumn{1}{c}{\textbf{Stat}} &
  \multicolumn{1}{c}{\textbf{P-Value}} &
  \multicolumn{1}{c}{\textbf{Stat}} &
  \multicolumn{1}{c}{\textbf{P-Value}} &
  \multicolumn{1}{c}{\textbf{Stat}} &
  \multicolumn{1}{c}{\textbf{P-Value}} &
  \multicolumn{1}{c}{\textbf{Stat}} &
  \multicolumn{1}{c}{\textbf{P-Value}} \\ \midrule
Restaurants &
  0,455 &
  2,2E-71 &
  0,682 &
  6,4E-189 &
  0,386 &
  4,7E-50 &
  \textbf{0,450} &
  1,5E-69 \\
Shop &
  0,480 &
  3,3E-80 &
  0,631 &
  2,0E-153 &
  0,449 &
  4,1E-69 &
  \textbf{0,501} &
  4,0E-88 \\
Officials &
  0,280 &
  3,0E-26 &
  0,342 &
  4,1E-39 &
  0,348 &
  1,6E-40 &
  \textbf{0,238} &
  3,5E-19 \\
Education &
  0,278 &
  6,9E-26 &
  0,392 &
  7,9E-52 &
  \textbf{0,266} &
  1,1E-23 &
  0,210 &
  4,0E-15 \\
Health &
  0,258 &
  2,5E-22 &
  0,406 &
  8,6E-56 &
  0,196 &
  2,1E-13 &
  \textbf{0,223} &
  5,8E-17 \\
Business &
  0,542 &
  1,1E-105 &
  0,739 &
  1,9E-237 &
  \textbf{0,508} &
  4,3E-91 &
  0,652 &
  2,5E-167 \\
Recreation &
  0,346 &
  5,4E-40 &
  0,507 &
  9,4E-91 &
  0,330 &
  3,3E-36 &
  \textbf{0,331} &
  1,6E-36 \\ \midrule
\multicolumn{1}{c}{\textbf{Total}} &
  0,675 &
  1,4E-183 &
  0,894 &
  0,0E+00 &
  0,637 &
  2,5E-157 &
  \textbf{0,713} &
  1,4E-213 \\ \bottomrule
\end{tabular}%
}
\end{table}

\subsection{Individual Attributes}
\begin{figure}[ht]
\centering
{{\includegraphics[scale=0.5]{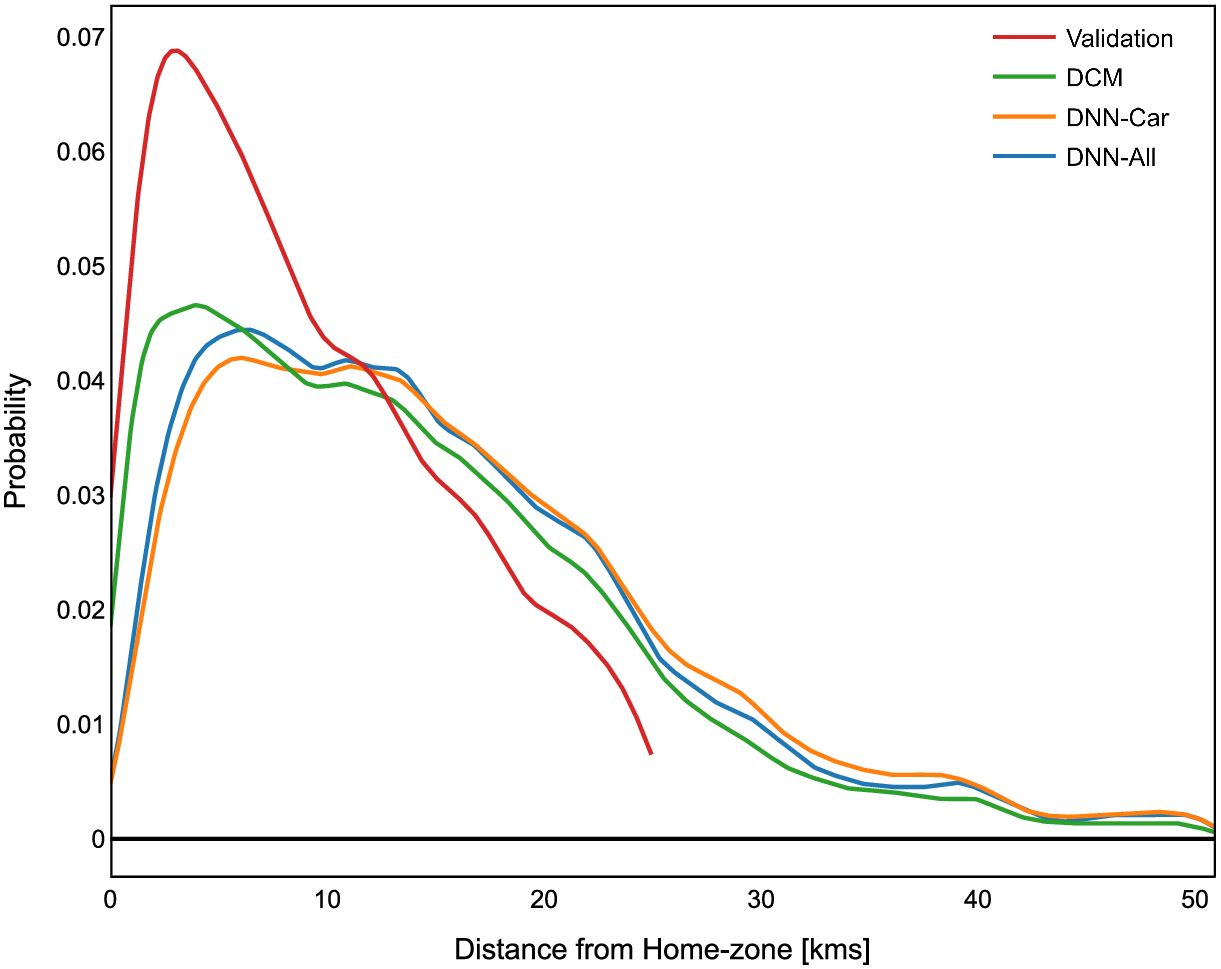}}}
\caption{Probability distribution of distance of chosen work from given home.}
\label{fig:DistProb}
\end{figure}

Commuting time is one of the main determinants of workplace location choice. Generally, individuals tend to select workplaces closer to their homes—a trend evident from the probability distributions of the distances between chosen work zones and corresponding home zones. Figure \ref{fig:DistProb} displays these distributions for the validation dataset as well as for the DCM and DNN models. For each model, the distances were computed by sampling 100 individuals from each zone, with the sampling weighted by each zone’s probability distribution for each individual.

In addition, we employ the two-sample Kolmogorov-Smirnov (KS) test to compare the probability distributions of distances from the chosen work zones to the home zones with that of the validation dataset. The KS test’s null hypothesis states that the two distributions are identical, while the alternative hypothesis suggests they are different. We use a 95\% confidence level, meaning that if the p-value is less than 0.05, the null hypothesis is rejected.

The plot shows that the distributions generated by the DNN and DCM models closely match the validation data, although some discrepancies exist. The DCM-based model produces a distribution that aligns more closely with the validation data at shorter distances and converges with the DNN-based models at longer distances. There is minimal difference between the distributions generated by the DNN-Car and DNN-All models, suggesting that incorporating additional individual attributes does not significantly improve distance-to-work predictions. These observations are supported by the KS test results in Table \ref{tab:ks-test}, where the DCM model exhibits lower KS statistics than the DNN models, indicating a better fit with the validation data, while the KS test statistics reveal negligible differences between the two DNN models.

\begin{table}[htbp]
\centering
\caption{KS statistics and p-value calculated by comparing distance of chosen workplaces by from each model with validation data.}
\label{tab:ks-test}
\resizebox{0.6\textwidth}{!}{%
\begin{tabular}{@{}cccc@{}}
           & \textbf{DCM} & \textbf{DNN-Car} & \textbf{DNN-All} \\ \midrule
Statistics & 0.177        & 0.256            & 0.246            \\
p-value    & 8.84E-127    & 1.02E-226        & 1.67e-218        \\ \bottomrule
\end{tabular}%
}
\end{table}

A fundamental assumption of the workplace choice model is that individual attributes significantly influence workplace zone selection. To visualize this effect, we examine the probability distributions of distances between chosen work zones and home zones, segmented by individual characteristics. In this study, we focus on two attributes: (1) gender and (2) access to a car. As shown in Table \ref{tab:ind-pearson}, these two attributes exhibit the strongest correlations with workplace choice in the validation data.

\begin{table}[htbp]
\centering
\caption{Pearson correlation coefficient and p-value between individual attributes and workplace chosen by individuals in the validation data.}
\label{tab:ind-pearson}
\resizebox{0.5\textwidth}{!}{%
\begin{tabular}{@{}lll@{}}
\multicolumn{1}{c}{\textbf{Attribute}} & \multicolumn{1}{c}{\textbf{Stat}} & \multicolumn{1}{c}{\textbf{P-Value}} \\ \midrule
gender           & \textbf{-0,077} & 2,5E-03 \\
employment       & -0,068          & 7,3E-03 \\
household income & 0,090           & 3,8E-04 \\
has kids?        & 0,120           & 2,1E-06 \\
household type   & 0,135           & 8,5E-08 \\
has car?         & \textbf{0,154}  & 1,1E-09 \\ \bottomrule
\end{tabular}%
}
\end{table}

\begin{figure}[htbp]
   \centering
   \begin{subfigure}{\textwidth}
        \centering
        \includegraphics[scale=0.4]{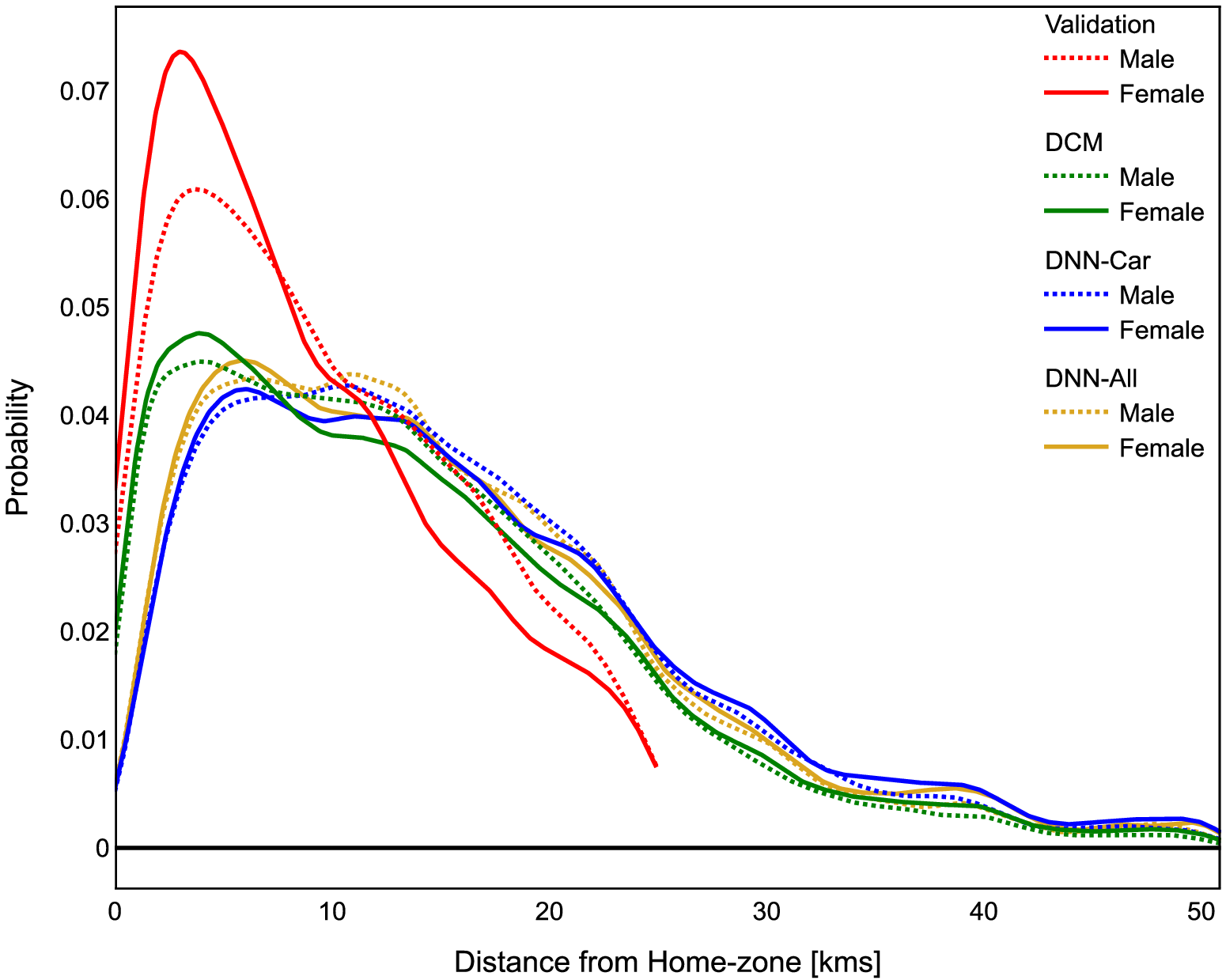}
        \caption{Gender}
        \label{fig:DistProb-sex}
    \end{subfigure}

    \begin{subfigure}{\textwidth}
        \centering
        \includegraphics[scale=0.4]{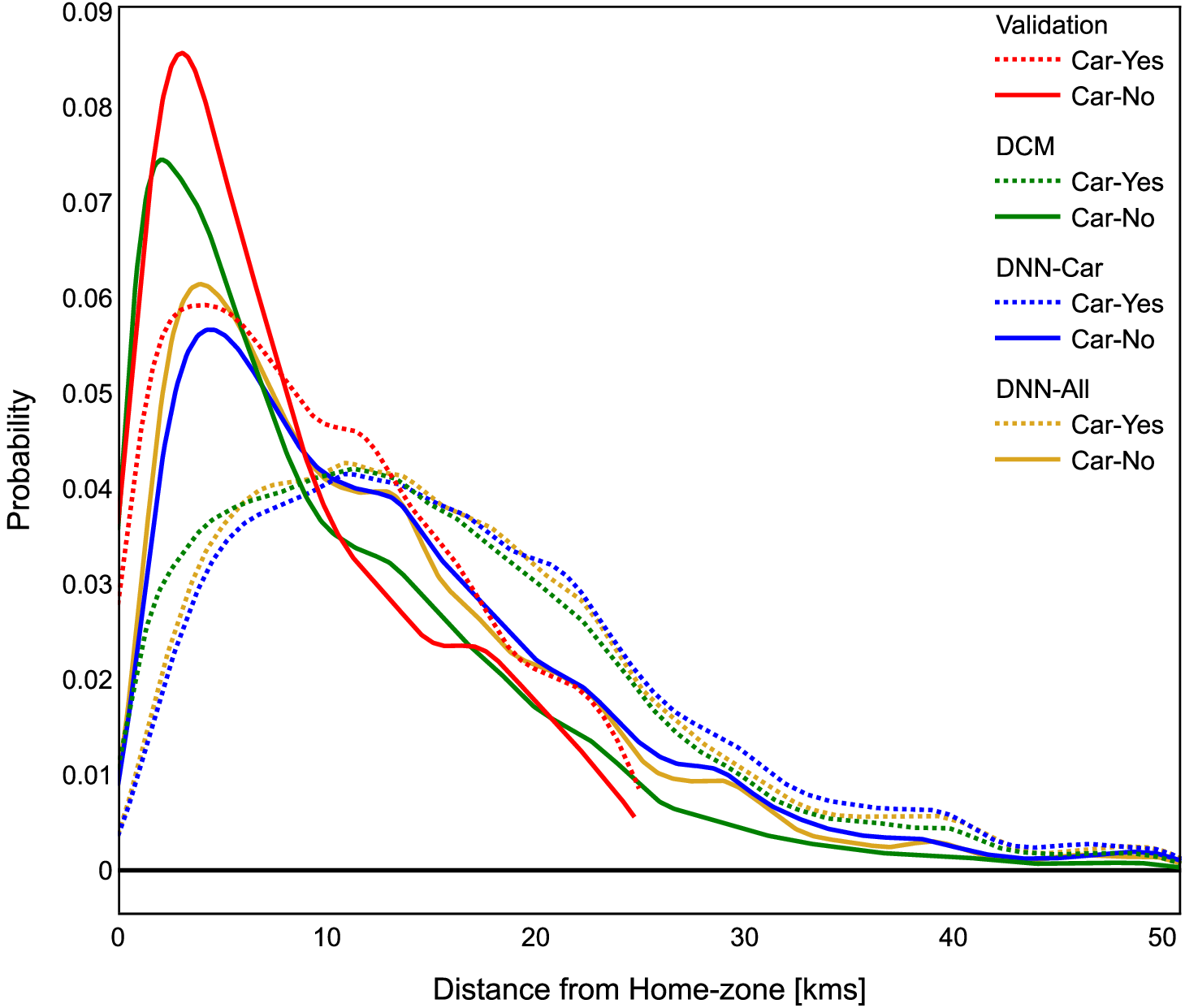}
        \caption{Access to Car}
        \label{fig:DistProb-car}
    \end{subfigure}

    \caption{Probability distribution of distance of chosen work from given home based on the individual attributes.}
\end{figure}

Figure \ref{fig:DistProb-sex} shows the probability distributions of distances between chosen work zones and home zones, segmented by gender, for the validation dataset and the outputs of both the DCM and DNN models. In the validation dataset, females tend to choose workplaces significantly closer to home than males. Although this trend appears in the model outputs, it is less pronounced, suggesting that both the DCM and DNN models may struggle to capture the subtle effect of gender on workplace choice. This limitation likely stems from the weak correlation between gender and workplace selection in the actual data, which poses a learning challenge for the models. The KS test results in Table \ref{tab:ks-sex} further support this observation—none of the models perfectly match the validation data, though the DCM model performs marginally better than the DNN models.

\begin{table}[htbp]
\centering
\caption{KS statistics and p-value calculated by comparing distance of chosen work-places by from each model with validation data, based on gender.}
\label{tab:ks-sex}
\resizebox{0.55\textwidth}{!}{%
\begin{tabular}{@{}cccc@{}}
           & \textbf{DCM} & \textbf{DNN-Car} & \textbf{DNN-All} \\ \midrule
\multicolumn{4}{c}{\textbf{Female}}                             \\ \midrule
Statistics & 0,204        & 0,287            & 0,263            \\
p-Value    & 1,47E-94     & 2,77E-188        & 6,86E-157        \\ \midrule
\multicolumn{4}{c}{\textbf{Male}}                               \\ \midrule
Statistics & 0,145        & 0,220            & 0,200            \\
p-Value    & 6,96E-38     & 5,71E-87         & 4,23E-72         \\ \bottomrule
\end{tabular}%
}
\end{table}

Similarly, Figure \ref{fig:DistProb-car} illustrates the probability distributions of distances between chosen work zones and home zones, this time segmented by car access, for the validation dataset and the outputs of the DCM and DNN models. In the validation data, individuals with access to a car are willing to travel farther for work compared to those without access. All models capture this trend, although the DCM model more effectively mirrors the distribution observed in the validation data. The KS test results in Table \ref{tab:ks-car} support this, showing that the DCM model has better statistics compared to the DNN models.

\begin{table}[htbp]
\centering
\caption{KS statistics and p-value calculated by comparing distance of chosen work-places by from each model with validation data, based on access to car.}
\label{tab:ks-car}
\resizebox{0.55\textwidth}{!}{%
\begin{tabular}{@{}cccc@{}}
           & \textbf{DCM} & \textbf{DNN-Car} & \textbf{DNN-All} \\ \midrule
\multicolumn{4}{c}{\textbf{Car - Yes}}                          \\ \midrule
Statistics & 0,220        & 0,282            & 0,259            \\
p-value    & 9,21E-133    & 3,78E-220        & 5,81E-185        \\ \midrule
\multicolumn{4}{c}{\textbf{Car - No}}                           \\ \midrule
Statistics & 0,092        & 0,235            & 0,207            \\
p-value    & 2,59E-11     & 1,12E-72         & 5,42E-56         \\ \bottomrule
\end{tabular}%
}
\end{table}

Our analysis indicates that while both the DCM and DNN models capture general trends in commuting behaviour, differences emerge in their ability to reflect nuances influenced by individual attributes. Specifically, although the DCM model appears to provide a slightly better fit for certain distance-related metrics, the choice of model may depend on which aspects of workplace choice are most critical to capture.

\newpage
\section{Conclusion}\label{Conclusion}
The primary aim of this research was to explore the application of DNNs in predicting workplace location choices and to compare their performance with traditional DCMs. This study aimed to use the flexibility and data-driven strengths of DNNs to overcome the weaknesses of DCMs, especially when dealing with complicated decisions that have many options.

This study thoroughly examines a DNN-based model for choosing workplace locations and a standard 2-level Nested logit discrete choice model (DCM). It indicates that both frameworks are statistically effective at predicting workplace locations while having their own unique benefits. A fully connected multi-layer feed-forward network with zone-specific utility blocks and alternative-specific constants was used to build the DNN model. It had higher log-likelihood values than the DCM benchmark. This is due to the DNN's ability to learn non-linear relationships and high-dimensional patterns on its own from input features without using pre-defined utility functions or nesting structures. By contrast, the DCM’s interpretable parameters provide clearer insights into behavioral mechanisms despite requiring explicit assumptions about decision hierarchy and variable interactions.

Furthermore, the attribute level comparison between the DNN and DCM models revealed that while both models are able to replicate the effect of the amount of job opportunities on workplace location choice, the DNN model performs better than the DCM. However, the DCM model demonstrated better alignment with the validation set when comparing the effect of individual attributes on the distance to the workplace. The DCM models work better for shorter distances, while DNN has comparable value with both validation and DCM for longer distances. This highlights the importance of selecting the appropriate model based on specific application needs.

In conclusion, this indicates that DNNs have a lot of potential as a strong tool for modeling workplace location choices, making them a suitable alternative to traditional DCMs. The findings suggest that DNNs can enhance the accuracy and robustness of predictions in transport science and urban planning, particularly in scenarios with complex decision-making processes. 

\subsection{Future Works}
The power of the DCM and the DNN models taken together may suggest intriguing hybrid future applications in ILUTMs. The DCM is good for simulating policies that need a precise estimate of elasticity. The DNN, on the other hand, will probably be more useful for short-term operations forecasts in transportation systems that are constantly changing. This could lead to more research into ensemble techniques that combine the pattern recognition power of a DNN with the behavioral approach of a DCM. These techniques could use joint latent space representations or attention-based feature selection processes. The advancement of methods should attempt to solve the problem of geographical generalization put on either model, thereby potentially overlooking inter-regional changes in urban structure and mobility cultures. Using graph neural networks to explicitly model the patterns of connections between zones could help make these results more useful in other situations.

\section{Acknowledgment}
The computations and data handling was enabled by the supercomputing resource Berzelius provided by National Supercomputer Centre at Linköping University and the Knut and Alice Wallenberg foundation. 

\bibliographystyle{apalike}
\bibliography{references}

\clearpage 
\end{document}